%% file: main.tex
\pdfoutput=1

\documentclass[11pt]{article}

\usepackage{acl}

\usepackage{todonotes}

\usepackage{times}
\usepackage{latexsym}

\usepackage[T1]{fontenc}

\usepackage[utf8]{inputenc}
\usepackage{graphicx}
\usepackage{microtype}

\usepackage{todonotes}

\title{Defending Compositionality in Emergent Languages}

\author{Michal Auersperger \and Pavel Pecina \\
  Charles University \\
  Faculty of Mathematics and Physics \\
  \texttt{\{auersperger,pecina\}@ufal.mff.cuni.cz}}

\begin{document}
\maketitle

\begin{abstract}
Compositionality has traditionally been understood as a major factor in productivity of language and, more broadly, human cognition. Yet, recently, some research started to question its status, showing that artificial neural networks are good at generalization even without noticeable compositional behavior.
We argue that some of these conclusions are too strong and/or incomplete.
In the context of a two-agent communication game, we show that compositionality indeed seems essential for successful generalization when the evaluation is done on a proper dataset.

\end{abstract}

\input{Intro}

\section*{Acknowledgements}
The work was supported by the grant 19-26934X (NEUREM3) by the Czech Science Foundation.

\bibliography{main}
\bibliographystyle{acl_natbib}

\appendix

\clearpage
\section{Appendix}
\label{sec:appendix}

\subsection{Architecture}
Both the sender and the receiver are implemented by the following encoder-decoder architecture:

The \textbf{encoder} produces two embeddings (size 500) for each input symbol, one syntactic and one semantic. A uni-directional GRU \citep{Chung14} layer (size 500) transforms the syntactic embeddings to contextualized embeddings.

The autoregressive \textbf{decoder} embeds the last produced symbol (or the start-of-sequence symbol) (size 500) and transforms it with another GRU layer (size 500). This contextualized embedding is then used as a query in the dot-product attention \citep{Luong15} and matched against the contextualized embeddings produced by the encoder. The attention weights are then used to produce the weighted sum of the semantic embeddings of the input symbols. This vector (size 500) is added to the query and transformed by a linear layer to the output symbol logits.

\end{document}

%% file: Intro.tex
\section{Introduction}
Compositionality is a property of language that describes its specific hierarchical structure. Multiple atomic units (e.g., words) can be combined to produce more complex units (e.g., sentences), while the meaning of the larger units can be inferred from the simpler parts and the way they are combined.

Of course, there are inherently noncompositional structures in language, idioms being a prime example. If someone is \textit{making waves}, it usually means that the person is causing trouble and no water is implied. Still, many reputable researchers \citep{Nowak00, Pinker00, Fodor02, Lake17} 
have seen compositionality as a key ingredient that enables language to be used productively, i.e., in an infinite number of novel situations. Often, this productivity is juxtaposed with the learning of Artificial Neural Networks (ANNs), whose performance is known to suffer when tested in new scenarios.

Recently, a strand in the literature has been making waves (\textit{sic}) by calling into question the proposed benefits of compositionality. Various papers have shown that ANNs can generalize well to unseen contexts (be productive) even if they work with internal representations that are noncompositional \citep{Kottur17, Andreas19, Baroni19, Chaabouni20, Kharitonov20}. 

The goal of this paper is to counterbalance these claims.\footnote{
We consider the cited research valuable and important. In some cases we argue with some statements that were not even the main topic of the paper in question.} More precisely, we would like to relativize some of the stronger conclusions that are provided and show (by running modified experiments) that the reported experimental results can be harmonized with the view that compositionality is necessary for successful knowledge transfer.

We will do this by first providing an overview of the related research in Section \ref{sec:summary_wrong}. Then in Section \ref{sec:our_criticism}, we give our arguments against some of the presented assumptions and/or conclusions. These arguments are further supported by experiments that are described in Section \ref{sec:experiments}. We conclude the paper with a short discussion in Section \ref{sec:discussion}.

\section{Emergent languages generalize without compositionality}
\label{sec:summary_wrong}
Much of the critical research comes from experimentation with languages that emerge during a communication game. Here, two agents (ANNs) are trained to communicate in order to perform a certain task. The first agent (sender/speaker) encodes the input into a message, a sequence of discrete symbols. The second agent (receiver/listener) does not have direct access to the original input, but only sees it as represented by the message. The receiver's goal is to transform the message into a desired output.

The output can take many forms depending on the task: reconstructing the input fully \citep{Chaabouni20, Andreas20} or partially \citep{Kottur17} or reconstructing the input after going through some deterministic transformation \citep{Kharitonov20}.

Inputs can be conceptualized as representations of objects by means of independent nominal attributes, e.g., \textit{blue circle, red square, orange triangle} for two-dimensional input vectors (color and shape). A uniform random sample of all such objects is then held out for testing, the rest of the data (or its part) is used for training. 


\citet{Andreas19} and \citet{Chaabouni20} develop custom metrics to measure the compositionality of messages passed between the sender and the receiver. The metrics compare each message to the corresponding input and try to assess to what extent each part of the input (attribute value) can be isolated in the message regardless of the context (i.e., other attribute values). They report runs in which models communicate through messages with low compositionality scores, but still achieve good generalization on unseen data.

\citet{Kharitonov20} replace training the first agent (sender) by hand-coding the messages. This gives them the advantage of having direct control over the emergent language. They find that good generalization can sometimes be achieved using a non-compositional language (sometimes leading to even better results than using a compositional language).

\section{Allowing compositionality to have an effect}
\label{sec:our_criticism}
Our main argument is that the process of selecting the test data plays a crucial role in evaluating the effects of compositionality. Sampling examples with uniform probability is a mainstay in machine learning, and algorithms have been shown many times that they can perform very well under these conditions. However, once we move away from the static world of i.i.d. data samples into the dynamic world of ever-changing distributions, the limitations of such models become obvious.

We argue that this is where compositionality is supposed to be helpful. Analyzing the world (or data points to keep the discussion down to earth) through a hierarchy of parts and their relations enables inferring a `rule-based algebraic system', which is `an extremely powerful generalization mechanism' \citep{Baroni19}. Systematic compositionality exploited by human learners enables them to be sample efficient, i.e., quickly learn a new task seeing just one or a limited number of training examples \citep{Lake17}. Therefore, we conclude that showing compositionality not being correlated with generalization on in-domain held-out data is not very informative. Instead, it is preferable to control the exposition of certain patterns in the training and testing data as illustrated by, e.g., the SCAN benchmark \citep{Lake18}.

We also point out that \citet{Andreas19} and \citet{Chaabouni20} implement both agents (sender and receiver) as relatively standard encoder/decoder architectures with recurrent (LSTM, GRU) layers. These architectures are not necessarily known for their ability to produce or utilize compositional sequences.

The assumption seems to be that when running an experiment many times (with different random initialization of the models), some runs will be successful in the sense that the agents will develop a more or less compositional language by chance. Moreover, the degree of compositionality would have to be large enough to influence the generalization of the models (should such an effect be real). We do not think this assumption is justified.

\citet{Kharitonov20} avoid a part of the above problem by creating compositional messages manually.
Their main conclusion is that one can devise different tasks, and in some of them a compositional representation of the input data might even prove disadvantageous. Indeed, it is possible to create, for example, an arbitrary bijective function from the input space to the output space and train a model to learn such a mapping. Not surprisingly, such a model will fail at the test time. However, if we first encode the input with the same arbitrary transformation (thus creating a non-compositional representation of the input data), the model is then asked to learn the identity function, which it might achieve quite well. Therefore, we agree with the conclusion that `in isolation from the target task, there
is nothing special about a language being \ldots compositional.'

\begin{table*}[t]
\renewcommand{\arraystretch}{1.2}
\small
\centering
\begin{tabular}{
    lccc
}
\textbf{Model} & \textbf{train} & \textbf{in-domain test} & \textbf{out-of-domain test}\\
\hline 
sender (ours)                & $1.00 \pm 0.01$ & $1.00 \pm 0.00$ & $\textbf{0.99} \pm \textbf{0.06}$ \\
sender \citep{Chaabouni20}   & $1.00 \pm 0.04$ & $1.00 \pm 0.03$ & $0.83 \pm 0.19$ \\
\hline
receiver (ours)              & $1.00 \pm 0.00$ & $1.00 \pm 0.00$ & $\textbf{1.00} \pm \textbf{0.02}$ \\
receiver \citep{Chaabouni20} & $1.00 \pm 0.02$ & $1.00 \pm 0.00$ & $0.44 \pm 0.32$ \\
\hline
\end{tabular}
\caption{\label{tab:partial-results} Learning alone experiment: Comparing the accuracy (mean $\pm$ std over 20 runs) of two types of architecture in different data splits. 
Each agent (sender/receiver) is trained independently of the other by using fixed messages.}
\end{table*}

Yet, as mentioned above, compositionality is often discussed in conjunction with natural language or human cognitive abilities more generally. In the lived experience of biological species (to be less human-centric), it is reasonable to expect that the `underlying factors of variation' or `explanatory factors' \citep{Bengio13} behind the input data a) repeat in many different situations and b) are directly relevant for different tasks. For example, it is reasonable to expect that many languages have a word for \textit{water} rather than a word for \textit{water and the wind is blowing}, simply because the first is more useful. Therefore, we believe that the emphasis on compositionality in research can be more than `a misguided effect of our human-centric bias' \citep{Kharitonov20}.

\section{Experiments}
\label{sec:experiments}
We follow the communication game experiments of \citet{Chaabouni20}. We create a set of instances, each of which is represented by $i_{att}$ attributes. Each attribute has $n_{val}$ possible values. The messages passed between the agents are limited by the maximum length ($c_{len}$) and the size of the vocabulary ($c_{voc}$). The receiver's goal is to reconstruct the input.

As messages are sequences of discrete symbols, which prevents gradients from passing through, the sender must be trained with the REINFORCE algorithm \citet{Williams92}. The receiver is trained using backpropagation. We use the EGG toolkit \citep{Kharitonov19} to implement the experiments.\footnote{The code is available at \url{https://github.com/michal-au/emlang-compos.git}}

We focus mainly on the setting of ($i_{att}$ = $2$, $n_{val}$ = $100$, $c_{len}$ = $3$, $c_{voc}$ = $100$). In this case, the dataset contains isntances such as $(12, 34), (0, 99), (99, 0)$ etc. We create three splits. The \textit{out-of-domain (OOD) test set} contains all pairs where 0 appears, apart from three examples: $(0, 0), (0, 1)$, and $(1, 0)$. The \textit{training set} contains these three zero examples together with 90\% of the remaining (nonzero) examples (random sample). The rest of the data constitute \textit{in-domain (IND) test set}.
In other words, we designate a special symbol ($0$), which appears only in a limited number of contexts in training. 
We then separately evaluate how the models perform on unseen examples with ordinary symbols (IND test set) and on unseen examples with the special symbol (OOD test set). 

Given the absence of any incentive for the models to develop compositional messages during training, we opt for architectural biases in our experiments. We use the models that have been proven to be successful in OOD generalization in the SCAN benchmark \citep{Li19, Russin20, Auersperger21}. 
Each agent is implemented as a separate seq2seq encoder-decoder architecture with recurrent layers and a modified attention mechanism. Details are provided in Appendix \ref{sec:appendix}


\subsection{Learning alone}
We first want to know whether the architectures used for implementing the agents are capable of achieving systematic compositionality on their own, that is, outside of the context of the 2-agent communication game. To do this, we handcode the messages that are to be created (sender) or received (receiver) and train each agent using regular backpropagation on the corresponding task.

We first create an arbitrary bijective mapping from the input vocabulary to the message vocabulary. Furthermore, to introduce variability in the length of the messages, we duplicate the occurrences of all tokens with odd indices in the message vocabulary. For example, having a mapping $tr: 0 \rightarrow 3, 1 \rightarrow 8, 2 \rightarrow 2$, \ldots, we produce (among others) the following input--message pairs: $(0, 0)$ -- $(3, 3)$, $(1, 0)$ -- $(8, 8, 3)$, $(2, 1)$ -- $(2, 2, 8, 8)$.

We train 2 types of agent for both sender and receiver comparing our architecture with the default used by \citet{Chaabouni20}. There are 500 training epochs and 20 runs with different random initializations for each agent. The results are presented in Table \ref{tab:partial-results}. They show that at least for this simple task, adding architectural bias helps with compositional generalization to out-of-domain data.
The results also suggest that the default architectures are unlikely to provide systematic generalization in the communication game since they are unable to achieve it even if their communication partner does never make mistakes (i.e. the messages are guaranteed to be produced/received correctly).

We experimented with limiting the capacity of the default architecture to see if such kind of regularization could help with generalization performance. Besides the original size (500), four additional sizes of hidden layers were tested (100, 200, 300, 400). For the default receiver, a smaller capacity (100) improved the OOD accuracy from $0.44$ ($\pm 0.32$ std) to $0.81$ ($\pm 0.21$ std).

We point out that it is the out-of-domain test set that reveals the difference between the architectures.

\subsection{Communication game}

\begin{figure}[t]
  \centering
  \includegraphics*[scale=0.3]{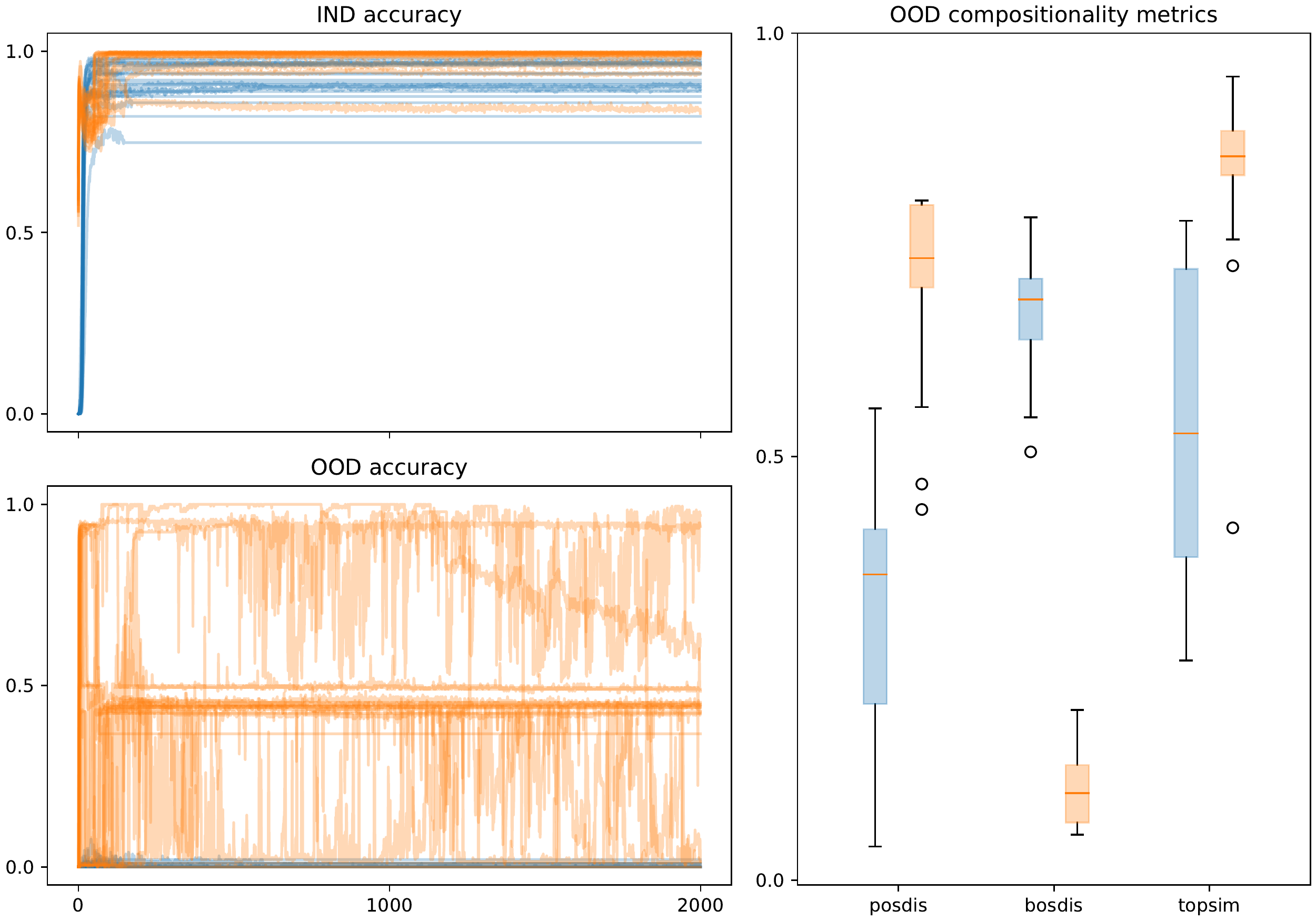}
  \caption{
    Communication game experiment: Training and OOD accuracy during training. Compositionality measures at the end of training.
    Orange represents runs of our architecture, blue represents runs of the architecture used by \citet{Chaabouni20}. There were 20 runs for each architecture.}
  \label{fig:full-training}
\end{figure}

Having seen that in some tasks our architectures are capable of approaching systematic compositionality, we turn our attention to the full communication game. We train both the default and our modified architectures on the full task for the maximum number of 2,000 epochs. Similarly to \citet{Chaabouni20}, we use early stopping when training accuracy reaches 99.999\%, however, we evaluate all the runs, even those that never reach perfect training accuracy. Each experiment was repeated 20 times with different random initializations. The training progress is visualized in Figure \ref{fig:full-training} and the results are given in Table \ref{tab:full-results}.

The experiments demonstrate that our changes to the default architecture lead to some out-of-domain generalization, but we were unable to guarantee such behavior for each run (accuracy $0.42 \pm 0.27$ std). In contrast, the original architecture does never succeed (accuracy $0.00 \pm 0.01$ std).

\begin{figure}[htb]
  \centering
  \includegraphics*[scale=0.3]{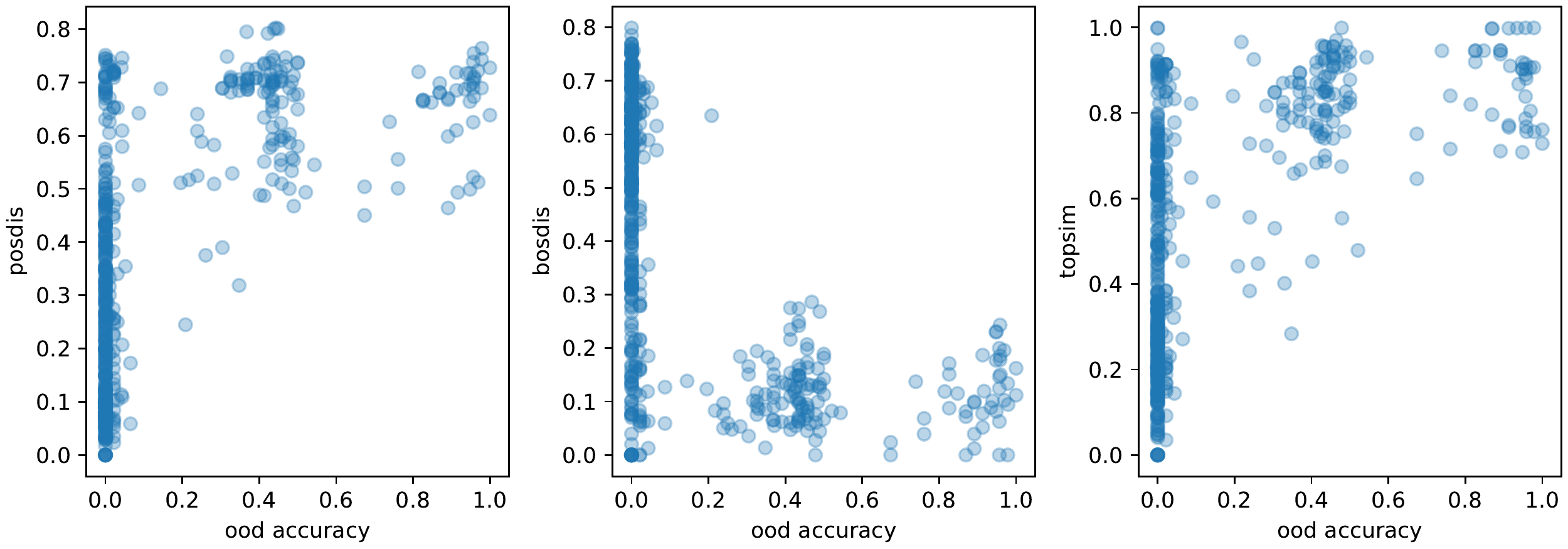}
  \caption{
    Compositionality measures and generalization in out-of-domain dataset. Successful generalization was possible only with large posdis and/or topsim scores. 
  }
  \label{fig:comp-vs-gen}
\end{figure}

We evaluated the three compositionality metrics used by \citet{Chaabouni20} and found significant differences between the two architectures. In two of the metrics, namely \textit{positional disentanglement} and \textit{topographic similarity}, we achieve higher scores than the original architecture, while in \textit{bag-of-symbol disentanglement} the situation is reversed. We show the relationship between compositionality and generalization in Figure \ref{fig:comp-vs-gen}. 
\begin{table*}[htb]
\renewcommand{\arraystretch}{1.2}
\small
\centering
\begin{tabular}{l|ccc|ccc}
\textbf{Model} & \textbf{train} & \textbf{IND test} & \textbf{OOD test} & \textbf{posdis} & \textbf{bosdis} & \textbf{topsim}\\
\hline
\citet{Chaabouni20}   & $0.99 \pm 0.07$ & $0.91 \pm 0.09$ & $0.00 \pm 0.01$ & $0.32 \pm 0.15$ & $0.67 \pm 0.07$ & $0.53 \pm 0.18$ \\
ours                  & $0.99 \pm 0.03$ & $0.98 \pm 0.05$ & $\textbf{0.42} \pm \textbf{0.27}$ & $0.71 \pm 0.11$ & $0.11 \pm 0.04$ & $0.83 \pm 0.11$ \\
\hline
\end{tabular}
\caption{\label{tab:full-results} Communication game experiment: Accuracy (mean $\pm$ std) measured in different data splits, Three compositionality measures of the messages (mean $\pm$ std) evaluated in out-of-domain test data. }
\end{table*}

Both \citet{Andreas19} and \citet{Chaabouni20} claim that compositionality is not a necessary condition for good generalization, but that it might be a sufficient condition \citep{Chaabouni20}. Choosing the out-of-domain data for evaluation and training models whose architecture is biased towards utilizing composationality of a language, we arrive at the opposite conclusion: compositionality is a necessary but not sufficient condition for good generalization. In other words, we often observe runs where both agents communicate through relatively compositional messages, but fail to generalize. However, we never observe a run where generalization is successful in spite of a low compositionality (posdis or topsim) score.

However, given the size of the input space ($100\!\times\!100$) and the proportion of training data (about $90\%$), we did not expect to find such a noticeable difference in performance in the in-domain test set. This suggests that such test data is not completely agnostic to the notion of compositionality (which is somewhat contrary to our previous argumentation). Yet, we still maintain that the out-of-domain dataset is much more informative with respect to evaluating the benefits of compositionality.

Similarly to the previous experiment, we tested additional sizes of hidden layers of the original architecture (100, 200, 300, 400) but were not able to match the IND accuracy of our architecture.

Looking at the OOD generalization performance of our models, it is notable that most models operate in one of three regimes: the accuracies tend to cluster above 0, below 0.5, or below 1. Manual inspection of the agents' communication showed that most of the time agents successfully reconstruct the non-zero symbols, which means that most of the errors are caused by wrongly reconstructing the zero symbol. These errors are also systematic, meaning that, given the position in the string, the zero symbol is replaced by the same symbol regardless of its neighbor. Thus, agents successfully reconstructing zero at both positions achieve accuracies close to 1, agents successfully reconstructing zero only in a single position achieve scores close to 0.5 and in the rest of the runs, agents fail regardless of the position.

\section{Discussion}
\label{sec:discussion}
There are many questions that remain for further analysis. The distinction between the in-domain and out-of-domain data is not clear-cut. One might object that seeing just one or two examples of a given symbol in the training data is too little for ANNs to learn its embedding and reliably map it close to other `similar' symbols in the semantic space \citep{Lake18, Loula18}. This is actually the issue as it seems that human learners unlike ANNs are able to succeed in such a scenario and work with limited data or, in other words, `not-yet-converged embeddings'. See \citet{Lake17} for a more thorough discussion. 

The goal of this paper was to show that some conclusions in the literature on compositionality are too strong or incomplete. However, there are other arguments that remain untackled. \citet{Baroni19} gives examples of neural networks that generalize (partially) well to out-of-domain data. For instance, \citet{Dessi19} show that a simple convolutional network is enough to improve accuracy from 1.2\% to 60\% in a difficult task from the SCAN benchmark. \citet{Gulordava18} demonstrate that a language model is capable of preferring grammatical nonsense sentences (certainly not seen in training) to ungrammatical ones. In general, the practical success of ANNs in many applications can serve as a proof of their strong generalization abilities \citep{Lake18, Baroni19}.

In response, we would like to point out that such success often coincides with new developments in neural architectures (convolutional NNs in vision, attention in NLP). These developments might actually point in the direction of compositionality. A trained convolutional NN actually detects primitive shapes (at least by the filters in the lower layers) and combines these into composit representations. Similarly, a trained attention-based encoder-decoder language model represents each input as a sequence of contextualized embeddings of the original units. Some of these embeddings might primarily represent the corresponding input units, and others might represent their collections.

We also acknowledge that for practical applications, especially in the short term, focusing on compositionality is not guaranteed to help \cite{Kharitonov20}. The most practical way so far has been to enable training on as much data as possible. However, it is likely that such an approach will eventually result in diminishing returns.

There are many potential aspects that favor compositional behavior: inductive architectural biases (e.g., attention); limited channel capacity relative to the input space \citep{Nowak00}; ease of transmission within the population \citep{Chaabouni20}; generalization pressure (contrary to \citealt{Chaabouni20}), if such a pressure is allowed to have some effect (e.g., by meta-learning); \ldots. It is also possible that systematic compositionality might emerge\footnote{As is often the case, \textit{emerge} here means \textit{somehow come about} and is used to hide the fact that the authors cannot explain things any further.
}
only as a result of multiple such factors.

%% file: main.bbl
\begin{thebibliography}{22}
\expandafter\ifx\csname natexlab\endcsname\relax\def\natexlab#1{#1}\fi

\bibitem[{Andreas(2019)}]{Andreas19}
Jacob Andreas. 2019.
\newblock \href {https://openreview.net/forum?id=HJz05o0qK7} {Measuring
  compositionality in representation learning}.
\newblock In \emph{International Conference on Learning Representations}.

\bibitem[{Andreas(2020)}]{Andreas20}
Jacob Andreas. 2020.
\newblock \href {https://doi.org/10.18653/v1/2020.acl-main.676} {Good-enough
  compositional data augmentation}.
\newblock In \emph{Proceedings of the 58th Annual Meeting of the Association
  for Computational Linguistics}, pages 7556--7566, Online. Association for
  Computational Linguistics.

\bibitem[{Auersperger and Pecina(2021)}]{Auersperger21}
Michal Auersperger and Pavel Pecina. 2021.
\newblock \href {https://aclanthology.org/2021.ranlp-1.11} {Solving {SCAN}
  tasks with data augmentation and input embeddings}.
\newblock In \emph{Proceedings of the International Conference on Recent
  Advances in Natural Language Processing (RANLP 2021)}, pages 86--91, Held
  Online. INCOMA Ltd.

\bibitem[{Baroni(2019)}]{Baroni19}
Marco Baroni. 2019.
\newblock \href {https://doi.org/10.1098/rstb.2019.0307} {Linguistic
  generalization and compositionality in modern artificial neural networks}.
\newblock \emph{Philosophical Transactions of the Royal Society B: Biological
  Sciences}, 375(1791):20190307.

\bibitem[{Bengio et~al.(2013)Bengio, Courville, and Vincent}]{Bengio13}
Yoshua Bengio, Aaron Courville, and Pascal Vincent. 2013.
\newblock Representation learning: A review and new perspectives.
\newblock \emph{IEEE Transactions on Pattern Analysis and Machine
  Intelligence}, 35(8):1798--1828.

\bibitem[{Chaabouni et~al.(2020)Chaabouni, Kharitonov, Bouchacourt, Dupoux, and
  Baroni}]{Chaabouni20}
Rahma Chaabouni, Eugene Kharitonov, Diane Bouchacourt, Emmanuel Dupoux, and
  Marco Baroni. 2020.
\newblock \href {https://doi.org/10.18653/v1/2020.acl-main.407}
  {Compositionality and generalization in emergent languages}.
\newblock In \emph{Proceedings of the 58th Annual Meeting of the Association
  for Computational Linguistics}, pages 4427--4442, Online. Association for
  Computational Linguistics.

\bibitem[{Chung et~al.(2014)Chung, Gulcehre, Cho, and Bengio}]{Chung14}
Junyoung Chung, Caglar Gulcehre, Kyunghyun Cho, and Yoshua Bengio. 2014.
\newblock Empirical evaluation of gated recurrent neural networks on sequence
  modeling.
\newblock In \emph{NIPS 2014 Workshop on Deep Learning, December 2014}.

\bibitem[{Dessi and Baroni(2019)}]{Dessi19}
Roberto Dessi and Marco Baroni. 2019.
\newblock \href {https://doi.org/10.18653/v1/P19-1381} {{CNN}s found to jump
  around more skillfully than {RNN}s: Compositional generalization in seq2seq
  convolutional networks}.
\newblock In \emph{Proceedings of the 57th Annual Meeting of the Association
  for Computational Linguistics}, pages 3919--3923, Florence, Italy.
  Association for Computational Linguistics.

\bibitem[{Fodor and Lepore(2002)}]{Fodor02}
Jerry~A. Fodor and Ernest Lepore. 2002.
\newblock \emph{Compositionality Papers}.
\newblock Oxford University Press UK.

\bibitem[{Gulordava et~al.(2018)Gulordava, Bojanowski, Grave, Linzen, and
  Baroni}]{Gulordava18}
Kristina Gulordava, Piotr Bojanowski, Edouard Grave, Tal Linzen, and Marco
  Baroni. 2018.
\newblock \href {https://doi.org/10.18653/v1/N18-1108} {Colorless green
  recurrent networks dream hierarchically}.
\newblock In \emph{Proceedings of the 2018 Conference of the North {A}merican
  Chapter of the Association for Computational Linguistics: Human Language
  Technologies, Volume 1 (Long Papers)}, pages 1195--1205, New Orleans,
  Louisiana. Association for Computational Linguistics.

\bibitem[{Kharitonov and Baroni(2020)}]{Kharitonov20}
Eugene Kharitonov and Marco Baroni. 2020.
\newblock \href {https://doi.org/10.18653/v1/2020.blackboxnlp-1.2} {Emergent
  language generalization and acquisition speed are not tied to
  compositionality}.
\newblock In \emph{Proceedings of the Third BlackboxNLP Workshop on Analyzing
  and Interpreting Neural Networks for NLP}, pages 11--15, Online. Association
  for Computational Linguistics.

\bibitem[{Kharitonov et~al.(2019)Kharitonov, Chaabouni, Bouchacourt, and
  Baroni}]{Kharitonov19}
Eugene Kharitonov, Rahma Chaabouni, Diane Bouchacourt, and Marco Baroni. 2019.
\newblock \href {https://doi.org/10.18653/v1/D19-3010} {{EGG}: a toolkit for
  research on emergence of lan{G}uage in games}.
\newblock In \emph{Proceedings of the 2019 Conference on Empirical Methods in
  Natural Language Processing and the 9th International Joint Conference on
  Natural Language Processing (EMNLP-IJCNLP): System Demonstrations}, pages
  55--60, Hong Kong, China. Association for Computational Linguistics.

\bibitem[{Kottur et~al.(2017)Kottur, Moura, Lee, and Batra}]{Kottur17}
Satwik Kottur, Jos{\'e} Moura, Stefan Lee, and Dhruv Batra. 2017.
\newblock \href {https://doi.org/10.18653/v1/D17-1321} {Natural language does
  not emerge {`}naturally{'} in multi-agent dialog}.
\newblock In \emph{Proceedings of the 2017 Conference on Empirical Methods in
  Natural Language Processing}, pages 2962--2967, Copenhagen, Denmark.
  Association for Computational Linguistics.

\bibitem[{Lake and Baroni(2018)}]{Lake18}
Brenden Lake and Marco Baroni. 2018.
\newblock \href {http://proceedings.mlr.press/v80/lake18a.html} {Generalization
  without systematicity: On the compositional skills of sequence-to-sequence
  recurrent networks}.
\newblock In \emph{Proceedings of the 35th International Conference on Machine
  Learning}, volume~80 of \emph{Proceedings of Machine Learning Research},
  pages 2873--2882, Stockholmsmässan, Stockholm Sweden. PMLR.

\bibitem[{Lake et~al.(2017)Lake, Ullman, Tenenbaum, and Gershman}]{Lake17}
Brenden~M. Lake, Tomer~D. Ullman, Joshua~B. Tenenbaum, and Samuel~J. Gershman.
  2017.
\newblock \href {https://doi.org/10.1017/S0140525X16001837} {Building machines
  that learn and think like people}.
\newblock \emph{Behavioral and Brain Sciences}, 40:e253.

\bibitem[{Li and Bowling(2019)}]{Li19}
Fushan Li and Michael Bowling. 2019.
\newblock \href
  {https://proceedings.neurips.cc/paper/2019/file/b0cf188d74589db9b23d5d277238a929-Paper.pdf}
  {Ease-of-teaching and language structure from emergent communication}.
\newblock In \emph{Advances in Neural Information Processing Systems},
  volume~32. Curran Associates, Inc.

\bibitem[{Loula et~al.(2018)Loula, Baroni, and Lake}]{Loula18}
Jo{\~a}o Loula, Marco Baroni, and Brenden Lake. 2018.
\newblock \href {https://doi.org/10.18653/v1/W18-5413} {Rearranging the
  familiar: Testing compositional generalization in recurrent networks}.
\newblock In \emph{Proceedings of the 2018 {EMNLP} Workshop {B}lackbox{NLP}:
  Analyzing and Interpreting Neural Networks for {NLP}}, pages 108--114,
  Brussels, Belgium. Association for Computational Linguistics.

\bibitem[{Luong et~al.(2015)Luong, Pham, and Manning}]{Luong15}
Thang Luong, Hieu Pham, and Christopher~D. Manning. 2015.
\newblock \href {https://doi.org/10.18653/v1/D15-1166} {Effective approaches to
  attention-based neural machine translation}.
\newblock In \emph{Proceedings of the 2015 Conference on Empirical Methods in
  Natural Language Processing}, pages 1412--1421, Lisbon, Portugal. Association
  for Computational Linguistics.

\bibitem[{Nowak et~al.(2000)Nowak, Plotkin, and Jansen}]{Nowak00}
Martin~A. Nowak, Joshua~B. Plotkin, and Vincent A.~A. Jansen. 2000.
\newblock \href {https://doi.org/10.1038/35006635} {The evolution of syntactic
  communication}.
\newblock \emph{Nature}, 404(6777):495--498.

\bibitem[{Pinker(2000)}]{Pinker00}
Steven Pinker. 2000.
\newblock \href {https://doi.org/10.1038/35006523} {Survival of the clearest}.
\newblock \emph{Nature}, 404(6777):441--442.

\bibitem[{Russin et~al.(2020)Russin, Jo, O{'}Reilly, and Bengio}]{Russin20}
Jacob Russin, Jason Jo, Randall O{'}Reilly, and Yoshua Bengio. 2020.
\newblock \href {https://doi.org/10.18653/v1/2020.acl-srw.42} {Compositional
  generalization by factorizing alignment and translation}.
\newblock In \emph{Proceedings of the 58th Annual Meeting of the Association
  for Computational Linguistics: Student Research Workshop}, pages 313--327,
  Online. Association for Computational Linguistics.

\bibitem[{Williams(1992)}]{Williams92}
Ronald~J. Williams. 1992.
\newblock \href {https://doi.org/10.1007/BF00992696} {Simple statistical
  gradient-following algorithms for connectionist reinforcement learning}.
\newblock \emph{Mach. Learn.}, 8(3–4):229–256.

\end{thebibliography}
